\documentclass{INTERSPEECH2023}
\usepackage{amsmath}
\usepackage{graphicx}
\usepackage{xcolor}

\usepackage{multirow}
\makeatletter
\def\thickhline{%
  \noalign{\ifnum0=`}\fi\hrule \@height \thickarrayrulewidth \futurelet
   \reserved@a\@xthickhline}
\def\@xthickhline{\ifx\reserved@a\thickhline`
               \vskip\doublerulesep
               \vskip-\thickarrayrulewidth
             \fi
      \ifnum0=`{\fi}}
\makeatother
\newlength{\thickarrayrulewidth}
\interspeechcameraready

\title{Multilingual Speech-to-Speech Translation into Multiple Target Languages}

\name{
  Hongyu Gong, Ning Dong, Sravya Popuri, Vedanuj Goswami, Ann Lee, Juan Pino
}
\address{Meta AI Research, USA}
\email{\{hygong, dnn, spopuri, vedanuj, annl, juancarabina\}@meta.com}

\begin{document}

\newcommand{\hongyu}[1]{{\color{orange}[\textbf{HG:}#1]}}
\newcommand{\jp}[1]{\textcolor{blue}{[JP: #1]}}

\maketitle
 
\begin{abstract}



Speech-to-speech translation (S2ST) enables spoken communication between people talking in different languages. Despite a few studies on multilingual S2ST, their focus is the multilinguality on the source side, i.e., the translation from multiple source languages to one target language. We present the first work on multilingual S2ST supporting multiple target languages. Leveraging recent advance in direct S2ST with speech-to-unit and vocoder, we equip these key components with multilingual capability. Speech-to-masked-unit (S2MU) is the multilingual extension of S2U, which applies masking to units which don't belong to the given target language to reduce the language interference. We also propose multilingual vocoder which is trained with language embedding and the auxiliary loss of language identification. On benchmark translation testsets, our proposed multilingual model shows superior performance than bilingual models in the translation from English into $16$ target languages.
\end{abstract}


\section{Introduction}

Speech-to-speech translation consists in translating an utterance from a source language into another language, preserving the semantic meaning. Traditional methods mostly build a pipeline of automatic speech recognition (ASR), machine translation (MT) and text-to-speech (TTS) synthesis \cite{DBLP:journals/taslp/NakamuraMNKKJZYSY06}. 
Recent research progress on direct approaches has paved the way for S2ST modeling without reliance on intermediate texts. Direct S2ST makes it possible to use only speech alignments as training data and support languages without standard writing systems \cite{DBLP:conf/icml/JiaRRP22}. The recently proposed direct approach uses discrete units learned from pre-trained HuBERT models as the bridge between source and target speech \cite{DBLP:conf/naacl/LeeGDSCWPAPGH22}. It builds a speech-to-unit (S2U) module to translate source speech to target units, and a separately trained vocoder constructs the target speech from these units.

Multilingual modeling has attracted great research interest in its scalability to the increased coverage of translation directions \cite{GoogleMT,DBLP:journals/corr/abs-2207-04672}.  
Instead of training and maintaining numerous bilingual models, we can use one multilingual model to support multiple directions. Besides deployment efficiency, the multilingual research is further motivated by the enhanced translation performance \cite{GoogleMT}. Translation is a resource-intensive task, however, not all languages have abundant training resources. The knowledge sharing is enabled by multilingual training across languages, benefiting a language with data in other languages. 

Research explorations have been made in multilingual speech-to-speech translation, but existing works focus on translation from multiple sources languages into English only \cite{DBLP:conf/lrec/JiaRWZ22,DBLP:journals/corr/abs-2211-04508}. To the best of our knowledge, this is the first study on multilingual S2ST supporting multiple target languages. We leverage the direct approach built upon S2U and vocoder, and further equip the model with the multilingual capability. Modeling challenges have been identified in order to support multiple target languages. First of all, languages have different unit vocabularies, and the concatenation of multiple unit sets increases the vocabulary size and makes the unit sequence modeling harder. Empirically we observe degraded translation performance with the extended unit dictionary. Secondly, monolingual vocoders used in existing S2ST studies do not scale efficiently with the increased language coverage in multilingual setting. 

In this work, we propose a speech-to-masked-unit model to address the first challenge of extended unit dictionary. We apply unit masking to help the model focus on the units belonging to the given language without being interfered by other languages. Another contribution of this work is the exploration of multilingual vocoders to synthesize speech for a family of similar languages. It effectively reduces the number of vocoders when scaling up target languages. To mitigate the language interference in multilingual speech synthesis, we add language embedding to vocoder training and introduce the auxiliary loss of language identification. Empirical results demonstrate positive transfer across languages in speech synthesis, and improved speech quality of multilingual vocoders. 

Our multilingual S2ST is empirically evaluated on the task of translating English into 16 languages. On the testsets from EuroParl \cite{DBLP:conf/icassp/Iranzo-SanchezS20}, VoxPopuli \cite{DBLP:conf/acl/WangRLWTHWPD20} and FLEURS \cite{DBLP:conf/slt/ConneauMKZADRRB22}, proposed multilingual models achieve consistent gains than bilingual models with an average of +$5.2$ and +$2.7$ BLEU on in-domain and out-of-domain data respectively. 

\section{Related work}


\textbf{Speech-to-speech translation}. Conventional approaches to S2ST are cascaded models with texts as intermediate outputs. Source speech is translated into target texts using speech-to-text translation or the combination of speech recognition and machine translation \cite{DBLP:journals/taslp/NakamuraMNKKJZYSY06}. Target texts are lastly converted to target speech via text-to-speech models. Direct S2ST models are recently proposed without the need of target texts. Translatotron 2 applies multitask learning with phoneme information \cite{DBLP:conf/icml/JiaRRP22}. Another type of direct models bridges source and target speech with units learned from acoustic models, and its framework consists of speech-to-unit and vocoder \cite{DBLP:conf/naacl/LeeGDSCWPAPGH22,DBLP:journals/corr/abs-2212-08055}. Besides the advances in translation modeling, recent works explore data mining \cite{DBLP:journals/corr/abs-2211-04508} and data augmentation \cite{nguyen2023improving} to improve the speech translation performance. 

\textbf{Multilingual modeling}. Multilinguality has been studied in machine translation \cite{GoogleMT}, automatic speech recognition \cite{DBLP:conf/interspeech/PratapSTHLSC20}, text-to-speech synthesis \cite{DBLP:conf/interspeech/NekvindaD20} and speech-to-text translation \cite{DBLP:conf/iwslt/TangGLWPSG21}. The advantages of multilingual models are the performance improvements brought by knowledge transfer across languages and better efficiency of model training and maintenance. Instead of training multiple monolingual models, researchers train a single multilingual model supporting numerous languages. As for speech-to-speech translation, a few recent works explore multilingual modeling from multiple source languages to one target language \cite{DBLP:conf/lrec/JiaRWZ22,DBLP:journals/corr/abs-2211-04508}. 

Despite positive transfer of cross-lingual knowledge, multilingual models are also faced with the challenge of language interference. It is known as the curse of multilinguality, which results in performance degradation in some language directions \cite{DBLP:conf/emnlp/WangLT20}. 




\section{Model}

To model speech-to-speech translation, we take advantage of the direct approach built upon speech-to-unit (S2U) and vocoder \cite{DBLP:journals/corr/abs-2211-04508,DBLP:conf/naacl/LeeGDSCWPAPGH22}. Given aligned source and target speech, the target speech is transformed into a sequence of discrete units with pre-trained HuBERT model \cite{DBLP:journals/taslp/HsuBTLSM21}. S2U model is trained to translate source speech to the corresponding target unit sequence. Vocoder is separately trained to synthesize speech from discrete units. In the stage of inference, units are predicted by S2U model from source speech, and then taken by the vocoder to synthesize target speech.

\begin{figure}[htbp!]
  \centering
  \includegraphics[width=\linewidth]{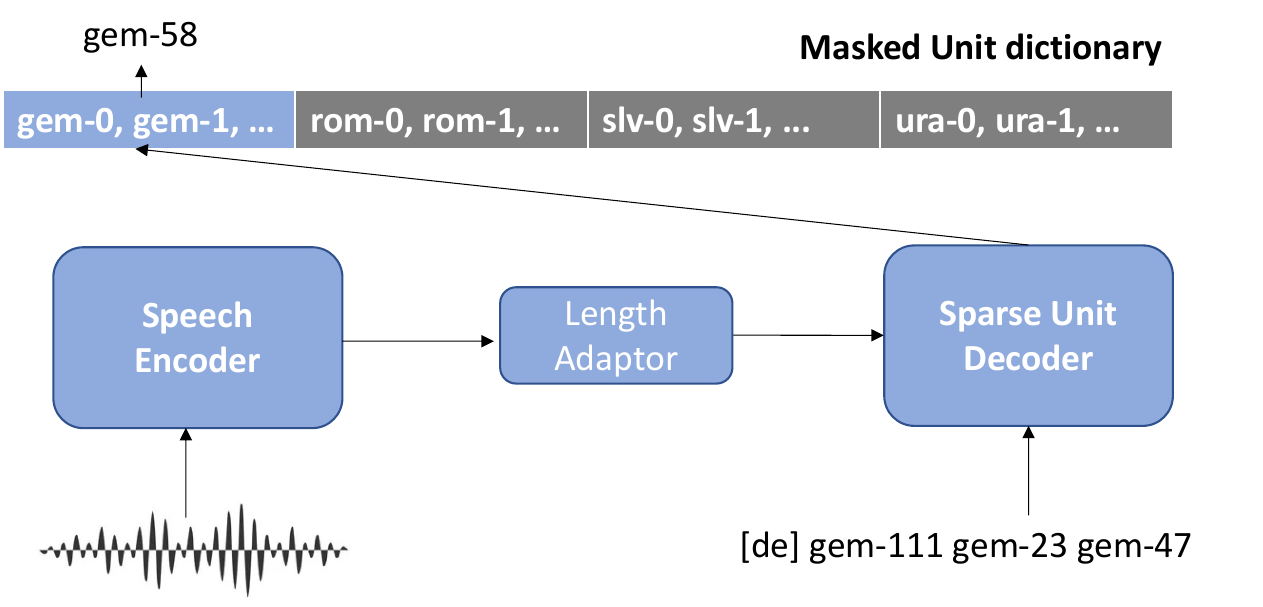}
  \caption{Architecture of multilingual speech-to-speech translation model.}
  \label{fig:s2s}
\end{figure}

Previous studies focus on only one target language in S2ST, and it is not trivial to adapt S2U model and vocoder to multilingual setting.
We propose a multilingual speech-to-masked-unit (S2MU) model as described in \autoref{subsec:s2mu}. 
Multilingual vocoders are introduced to improve speech synthesis quality across languages in \autoref{subsec:vocoder}.

\subsection{Speech-to-Masked-Unit Model}
\label{subsec:s2mu}

Our multilingual speech-to-masked-unit model has an encoder-decoder architecture. The overview of multilingual S2MU model is presented in \autoref{fig:s2s}. The speech encoder consists of convolutional layers and Transformer encoder layers, and the unit decoder is essentially a Transformer decoder. There is a length adaptor to bridge the sequence length gap between encoder outputs and decoder units. The adaptor is a single convolutional layer to downsample the encoder states since encoder length is longer than unit length. Similar to previous works \cite{DBLP:conf/interspeech/PopuriCWPAGHL22}, we initialize S2U model with pretrained encoder and decoder as the initialization demonstrated performance gains.

When supporting multiple target languages, the decoder needs language information to make correct predictions. Therefore we inform the decoder by prepending language tag to the target unit sequence. For example, ``[de]'' is prepended to the German units in \autoref{fig:s2s}. Suppose that multiple languages fall into language families such as Germanic (abbreviated as gem), Romanian (rom), Slavic (slv) and Uralic (ura) family. Each language family has their own unit dictionary, and languages within the same family share units since some of their pronunciations sound similar. To distinguish units in different vocabularies, we add the family tag to their units, i.e., Germanic units ``111 23 47'' are converted to ``gem-111, gem-23, gem-47''. Family dictionaries are then concatenated as extended target dictionary used by the unit decoder. 

The extended dictionary inevitably makes the unit prediction harder for the model, 
and units from other languages act as distractors in both training and inference. Empirically it is often seen that the model predicts units belonging to another language even when the target language is specified. We propose \textbf{unit masking}, which masks units from irrelevant languages in both decoder training and evaluation. It helps the model to focus on units in the target language no matter how large the extended unit vocabulary is. 

Suppose that the unit dictionary is $\mathbf{u}=\{u_{i}\}_{1\leq i\leq \left|V\right|}$ and $\left|V\right|$ is the vocabulary size. The index set of units in language $l$ is $\mathbf{m}^{l}$, i.e., $u_{i}$ belongs to language $l$ for $i \in \mathbf{m}^{l}$.
Denote $\mathbf{y}$ as ground truth target units, and $\hat{\mathbf{y}}$ as the predicted likelihood over units.
The training loss $L$ of speech-to-masked-unit model is calculated over only language $l$'s units instead of the whole unit dictionary.
\begin{align}
    L = -\frac{1}{T}\sum_{j=1}^{T}\sum_{i\in \mathbf{m}^{l}} y_{j,i}\log\hat{y}_{j,i}
\end{align}
where $y_{j,i}$ is a binary value indicating whether $u_{i}$ is the $j$-th unit in the target sequence and $T$ is the target length. 

As for inference, the predicted likelihood of units in other languages is forced to be $-\text{inf}$, so only units related to the target language are generated.

\subsection{Multilingual Vocoder}
\label{subsec:vocoder}

\begin{figure}[htbp!]
  \centering
  \includegraphics[width=\linewidth]{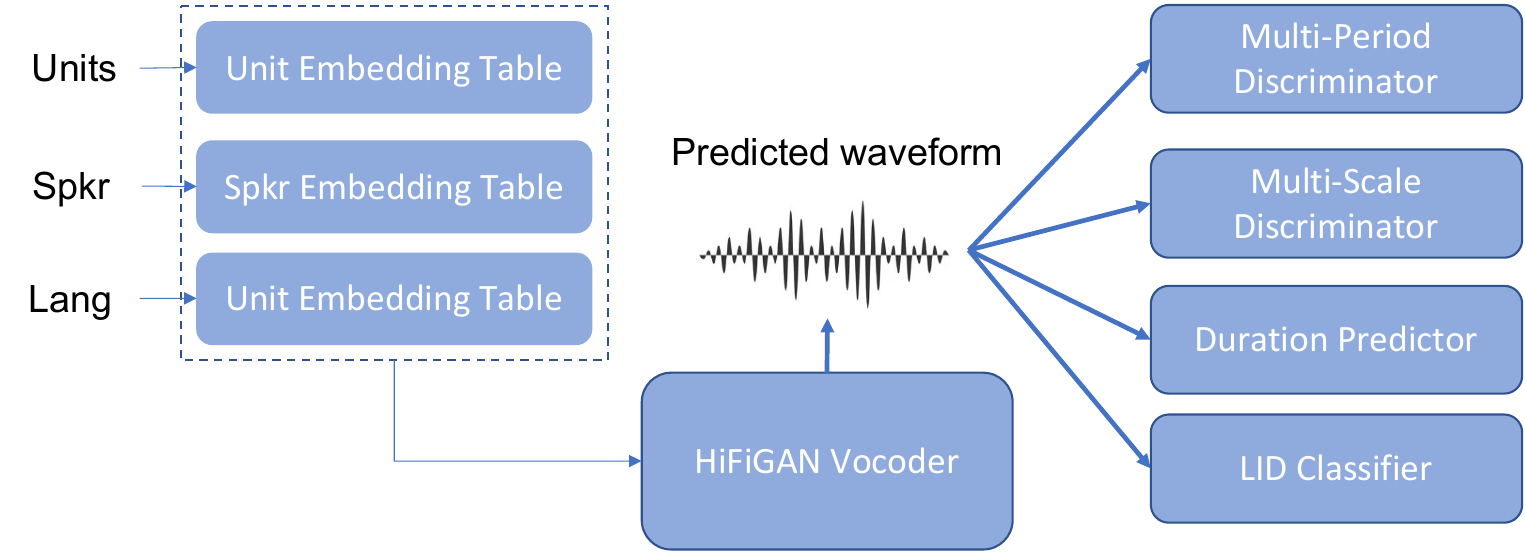}
  \caption{Architecture of multilingual vocoder.}
  \label{fig:mlg_vocoder}
\end{figure}

A monolingual vocoder typically consists of a HiFi-GAN generator which converts discrete units to speech waveform, a duration predictor and discriminators which provides feedback on the speech quality \cite{DBLP:conf/interspeech/PolyakACKLHMD21}. To extend it to multilingual setting, we introduce new components to the vocoder architecture as shown in Fig.~\ref{fig:mlg_vocoder}. Vocoders keep embedding tables to convert discrete units, speaker and language to continuous embeddings. We add the language tag to the input unit sequence, and an embedding lookup table retrieves language embedding and prepends it to the unit and speaker embedding. A common challenge of multilingual training is the language interference, and we notice that the generated speech from a multilingual vocoder might sound like another language. To mitigate the issue, we add a speech language identification (LID) classifier to the generator-discriminator framework. The LID classifier built on convolutional layers takes speech signal and predicts its language. 
Given input $\mathbf{x}$, the convolution layer consists of convolution operations followed by ReLU activation and LayerNorm.
\begin{align}
    \text{ConvLayer}(x) = \text{LayerNorm}(\text{ReLU}(\text{Conv}(\mathbf{x}))).
\end{align}

A linear projection layer is added on the top of LID classifier to predict the language of synthesized waveform. 
\begin{align}
    \hat{\mathbf{y}} = \text{softmax}(\mathbf{W}\cdot\text{ConvLayer}(\text{ConvLayer}(\mathbf{x}))),
\end{align}
where $\hat{\mathbf{y}}$ is predicted likelihood over languages, and $\mathbf{W}$ is a tunable weight matrix. The LID prediction indicates how well the generated speech fits in the given language. 

We outline how a multilingual vocoder trains generator together with auxiliary modules including LID classifier, duration predictor, Multi-Period (MPD) and Multi-Scale Discriminators (MSD). At each step, generator generates waveform based on discrete units together with speaker and language information.

\textbf{Auxiliary module training}. MPD and MSD are trained to distinguish the synthetic waveform from real speech. Duration predictor is tuned to predict the duration of consecutive units. Real speech is fed to LID classifier for language prediction.

\textbf{Generator training}. Generator is trained with multiple losses. The generated speech is compared with the reference via L1 loss of their mel-spectrograms and discriminator features. Adversarial loss is also applied to generator so that it learns to fool discriminators. Lastly LID classifier predicts the language of synthesized speech, and the LID loss penalizes generator for speech which does not sound like desired language.

\section{Experiments}

In the experiments, we focus on speech-to-speech translation from English into $16$ languages which are grouped into $4$ families based on their linguistic similarity. 
\begin{itemize}
    \item Germanic family: German (de) and Dutch (nl);
    \item Romance family: Spanish (es), French (fr), Italian (it), Portuguese (pt) and Romanian (ro);
    \item Slavic family: Czech (cs),  Croatian (hr), Lithuanian (lt), , Polish (pl),, Slovak (sk) and Slovenian (sl);
    \item Uralic family: Estonian (et), Finnish (fi) and Hungarian (hu).
\end{itemize}

\begin{table*}[htbp!]
\caption{WER ($\downarrow$) of resynthesized speech from vocoders (mono: monolingual vocoder, multi: multilingual vocoder with language embedding, multi (+LID): multilingual vocoder with language embedding and LID auxiliary loss), S and L indicate vocoder size as small or large.}
\label{tab:vocoder_wer}
\centering
\begin{tabular}{ccccccccc}
\thickhline 
\textbf{Family} & \textbf{Lang} & \textbf{Data} & \textbf{Train hours} & \textbf{ASR WER} & \textbf{Mono-S} & \textbf{Multi-S (+LID)} & \textbf{Multi-L} & \textbf{Multi-L (+LID)} \\ \thickhline
\multirow{2}{*}{\textbf{Gem}} & de & CSS10 & 13.2 & 10.0 & 16.1 & 13.1 & 14.2 & 13.1 \\
 & nl & CSS10 & 11.3 & 19.0 & 28.0 & 28.1 & 29.0 & 27.7 \\ \hline
\multirow{5}{*}{\textbf{Rom}} & es & CSS10 & 23.4 & 8.4 & 11.3 & 11.7 & 11.8 & 11.1 \\
 & fr & CSS10 & 17.7 & 24.0 & 30.7 & 30.2 & 28.7 & 28.6 \\
 & it & VoxPopuli & 25.8 & 23.0 & 31.6 & 32.5 & 29.6 & 28.7 \\
 & pt & Common Voice & 16.1 & 6.0 & 36.6 & 30.9 & 31.0 & 29.7 \\
 & ro & VoxpoPuli & 25.5 & 42.0 & 50.0 & 53.5 & 51.9 & 51.5 \\ \hline
\multirow{5}{*}{\textbf{Slv}} & cs & VoxPopuli & 26.8 & 15.0 & 23.0 & 24.2 & 24.4 & 23.0 \\
 & hr & VoxPopuli & 25.3 & 21.0 & 29.7 & 30.7 & 31.2 & 29.7 \\
 & lt & VoxPopuli & 1.3 & 38.0 & 57.3 & 57.3 & 57.3 & 57.3 \\
 & pl & VoxPopuli & 26.7 & 14.0 & 25.0 & 22.7 & 23.8 & 21.7 \\
 & sk & VoxPopuli & 25.3 & 28.0 & 41.0 & 40.8 & 41.7 & 38.9 \\
 & sl & VoxPopuli & 6.1 & 37.0 & 47.0 & 49.2 & 48.5 & 45.9 \\ \hline
\multirow{3}{*}{\textbf{Ura}}  & et & Common Voice & 12.0 & 14.0 & 44.1 & 47.5 & 47.9 & 45.9 \\
& fi & CSS10 & 8.3 & 2.0 & 17.8 & 18.7 & 17.7 & 16.4 \\
& hu & CSS10 & 7.9 & 21.0 & 21.0 & 28.8 & 28.0 & 24.9 \\ \thickhline
\end{tabular}
\end{table*}

The multilingual speech alignments are provided by SpeechMatrix \cite{DBLP:journals/corr/abs-2211-04508} together with useful resources including multilingual HuBERT models and vocoder training data.

\subsection{Empirical Setup}

\textbf{Preprocessing}. Speech-to-unit models and vocoders rely on units extracted with HuBERT and k-means models. Given speech alignments, we transform target speech into target units, and take the aligned source speech and target units as the S2U training data. As for vocoder training, we derive units from speech data, and vocoder is trained to reconstruct speech from the corresponding units.

We reuse multilingual HuBERT models provided by SpeechMatrix to learn speech features. Each HuBERT model was trained on audios collected from a family of languages, and thus speech features of languages from the same family are in the same space.  We further learn a k-means model for each family to cluster speech features. The continuous features are discretized by its cluster index assigned by the k-means model. Therefore a family of languages share the same unit vocabulary, and the number of clusters is its vocabulary size. The total vocabulary size of all languages is the sum of family vocabulary sizes.

To optimize the unit quality, previous works sweeped over multiple configurations of unit extraction. Following SpeechMatrix \cite{DBLP:journals/corr/abs-2211-04508}, we try different HuBERT layers (layer $10$, $11$ and $12$) for speech feature extraction and different cluster sizes for k-means clustering ($800$, $1000$, $1500$ and $2000$). In each configuration with a specific HuBERT layer and cluster size, we prepare a set of family units for vocoder training. Monolingual vocoders are trained on these family units and then evaluated on speech resynthesis as in \autoref{exp:vocoder}. The best configuration of unit extraction is selected based on the corresponding vocoder quality. In our experiments, we choose HuBERT layer $11$ for feature extraction in all languages, and the optimal k-means cluster sizes varies from family to familiy. Germanic, Slavic and Uralic families have the best cluster size of $1000$, Roman family has the best size of $2000$.

\textbf{Evaluation}. The performance of both vocoder and S2ST models is measured by the quality of their generated speech. As we care about the semantic preservation in the output speech, so we transcribe the speech into texts which carry the semantic content with pretrained ASR models. We reuse ASR models in \cite{DBLP:journals/corr/abs-2211-04508} for a fair comparison, which are built upon pretrained XLS-R or wav2vec2 models and finetuned on ASR datasets. These ASR models are released on HuggingFace \cite{huggingface}, and could be indexed by models ids as summarized in Appendix. The transcriptions of speech are lastly compared with reference texts, and different metrics are applied to measure how much they differ or resemble.

For vocoder evaluation, the metric is word error rate (WER) between speech transcriptions and ground truth texts. The lower WER, the better speech resynthesis a vocoder has. As for speech-to-speech translation, a commonly used metric is BLEU reflecting the lexical overlap. Higher BLEU score reflects better translation quality. 

We first evaluate the quality of multilingual vocoders on speech resynthesis in \autoref{exp:vocoder}. Next we train multilingual speech-to-masked-unit models, and report translation quality in comparison with bilingual models \autoref{exp:s2st}.

\subsection{Multilingual Vocoder}
\label{exp:vocoder}

\textbf{Dataset}.  We reuse the traininig and evaluation data as used by vocoders \footnote{We note that SpeechMatrix vocoders use units extracted from language-specific k-means models. In our experiments, both multilingual and monolingual vocoders use units from family-specific k-means model. Language- and family-specific units lead to comparable resynthesis quality for monolingual vocoders (c.f. Appendix)} in \cite{DBLP:journals/corr/abs-2211-04508}. Vocoder training requires high-quality speech which is collected from CSS10 \cite{DBLP:conf/interspeech/ParkM19}, VoxPopuli \cite{DBLP:conf/acl/WangRLWTHWPD20} and Common Voice \cite{DBLP:conf/lrec/ArdilaBDKMHMSTW20}. \autoref{tab:vocoder_wer} summarizes vocoder data statistics.

We develop multilingual vocoders for each language family, and combine all vocoder data in the same family as the train set. As a comparison, we also train monolingual vocoders for each language using the same set of speech and units. When it comes to evaluation, a trained vocoder takes test units and synthesizes speech which is then transcribed by pre-trained ASR models. We report word error rate of the transcriptions compared with reference texts in \autoref{tab:vocoder_wer}. 

\textbf{Hyperparameters}. The dimension of speaker embedding is set as $128$ for all vocoders. For multilingual vocoder, it has additional $128$-d language embedding.  The dimension of unit embeddings control the model capacity of vocoders, and we try a small architecture by setting unit dimension as $128$ and a large architecture by increasing unit dimension to $256$. 

The unit embeddings are upsampled by $5$ transposed convolutional layers to match the audio sample rate. The speaker embedding is concatenated with the upsampled representations and then processed by $15$ residual blocks which consist of dilated convolutional layers. For multilingual vocoder, when LID auxiliary loss is applied, we add an LID classifier which has two convolutional layers and a linear projection layer.

All vocoders are trained with a learning rate of $0.0002$ and a batch size of $16$. The training time does not differ much between monolingual and multilingual vocoders, and it takes around $3$ days on $8$ GPUs.



\textbf{Results}. In \autoref{tab:vocoder_wer}, we report WER of pretrained ASR models by providing the speech from the test set as the input and comparing ASR outputs with ground truth texts. We note that the WER metric is dependent on the ASR model quality, and ASR WER serves as a lower bound of vocoder WER. Therefore, it reflects the vocoder quality more accurately to check vocoder WER with respect to ASR WER. 
 
We compare vocoders of different model sizes and training recipes.  ``Mono-S'' and ``Multi-S'' are monolingual and multilingual vocoders that both have small architecture with $128$-d unit embeddings.  ``Multi-S'' prepends language id to vocoder inputs, and the training objectives are the same as monolingual vocoder ``Mono-S''. We also try larger architectures with $256$-d unit embeddings, i.e.,``Multi-L'' in \autoref{tab:vocoder_wer} ``Multi-S/L (+LID)'' are multilingual vocoders with the auxiliary LID loss as well as language embedding. 

Comparing ``Mono-S'' and ``Multi-S (+LID)'' which both have small architecture in \autoref{tab:vocoder_wer} , we find that multilingual model achieves comparable performance in Germanic, Romanian and Slavic family, and falls behind in Uralic languages. Multilingual performance can be further improved when we enlarge the architecture of ``Multi-S (+LID)'' to ``Multi-L (+LID)''.  The training with LID loss achieves lower WER than ``Multi-L'' without LID. The best vocoders across families are large multilingual vocoders trained with LID loss.


\subsection{English-to-Many S2ST}
\label{exp:s2st}

\textbf{S2ST data}. The multilingual dataset used in S2ST experiments is SpeechMatrix corpus with speech alignments between $17$ languages. We use parallel speech in 16 en-xx directions to train one-to-many S2ST models. SpeechMatrix is a mined corpus and each alignment is scored by its semantic similarity \cite{DBLP:journals/corr/abs-2211-04508}. We select aligned speech with scores above $1.09$ so that we could have a decent amount of good-quality training data. \autoref{tab:sm_data} reports the statistics of parallel speech data.

\begin{table}[htbp!]
\caption{Training data statistics of SpeechMatrix.}
\label{tab:sm_data}
\centering
\resizebox{0.45\textwidth}{!}{
\begin{tabular}{ccccccccc}
\thickhline
\textbf{Lang} & cs & de & es & et & fi & fr & hr & hu \\
\textbf{Hours} & 883 & 1,451 & 1,366 & 321 & 426 & 1,517 & 148 & 434 \\ \hline
\textbf{Lang} & it & lt & nl & pl & pt & ro & sk & sl \\
\textbf{Hours} & 1,575 & 4 & 1,231 & 942 & 988 & 521 & 593 & 46 \\ \thickhline
\end{tabular}}
\end{table}


\textbf{Models}. We implemented a multilingual speech-to-masked-unit model. The speech encoder is a stack of $7$ convolutional layers and $48$ Transformer encoder layers with $1024$-d and $4096$-d layer and forward embeddings. The unit decoder is a $12$-layer Transformer decoder with layer and forward dimensions of $1024$ and $4096$ respectively. A multilingual S2MU model has $1.2$B parameters. We initialize the speech encoder with XLS-R model of $1$B parameters \cite{DBLP:conf/interspeech/BabuWTLXGSPSPBC22} and initialize the unit decoder with mBART decoder trained on English units \cite{DBLP:conf/interspeech/PopuriCWPAGHL22}.

We include two bilingual approaches proposed in recent works as baselines. One is bilingual speech-to-unit (S2U) model \cite{DBLP:conf/interspeech/PopuriCWPAGHL22}, which has the same initialized speech encoder and unit decoder as the multilingual S2MU model. Its difference from S2MU lies in the decoder vocabulary. Since a bilingual S2U model supports one translation direction, the vocabulary of S2U only contains target language-specific units. A bilingual S2U model also has $1.2$B parameters.

\begin{table*}[htbp!]
\caption{ASR-BLEU scores of S2ST models on in-domain and out-of-domain testsets.}
\label{tab:s2s_bleu}
\centering
\resizebox{1.0\textwidth}{!}{
\begin{tabular}{cccccccccccccccccccc} \thickhline
\textbf{Domain} & \textbf{Model} & \textbf{Vocoder} & \textbf{cs} & \textbf{de} & \textbf{es} & \textbf{et} & \textbf{fi} & \textbf{fr} & \textbf{hr} & \textbf{hu} & \textbf{it} & \textbf{lt} & \textbf{nl} & \textbf{pl} & \textbf{pt} & \textbf{ro} & \textbf{sk} & \textbf{sl} & \textbf{avg} \\ \hline
\multicolumn{1}{c|}{\multirow{5}{*}{EP/VP}} & \multirow{2}{*}{S2MU} & Mono. & 10.5 & 15.5 & 23.1 & - & 2.6 & 19.2 & 2.6 & 1.1 & 15.0 & 0.1 & 18.6 & 10.1 & 12.3 & 8.8 & 1.2 & 4.3 & 9.7 \\
\multicolumn{1}{c|}{} &  & Multi. & 10.4 & 16.2 & 25.1 & - & 2.3 & 20.0 & 2.6 & 0.9 & 14.9 & 0.1 & 19.1 & 10.6 & 16.2 & 8.7 & 1.3 & 4.4 & 10.2 \\ \cline{2-20}

\multicolumn{1}{c|}{} & \multirow{2}{*}{S2U} & Mono. & 2.9 & 13.3 & 20.1 & - & 0.0 & 12.6 & 0.0 & 0.0 & 7.0 & 0.0 & 18.8 & 0.0 & 0.0 & 0.3 & 0.0 & 0.0 & 5.0 \\
\multicolumn{1}{c|}{} & & Multi. & 2.9 & 13.8 & 21.8 & - & 0.0 & 13.1 & 0.0 & 0.0 & 7.1 & 0.0 & 19.4 & 0.0 & 0.0 & 0.3 & 0.0 & 0.0 & 5.2 \\ \cline{2-20}
\multicolumn{1}{c|}{} & Textless & Mono. & 8.2 & 10.1 & 21.9 & - & 1.9 & 19.2 & 8.4 & 1.1 & 11.5 & 0.3 & 15.1 & 8.2 & 11.8 & 7.6 & 5.7 & 5.5 & 9.1 \\ \thickhline
\multicolumn{1}{c|}{\multirow{5}{*}{FLEURS}} & \multirow{2}{*}{S2MU} & Mono. & 4.3 & 6.4 & 7.8 & 1.5 & 0.9 & 12.4 & 2.9 & 0.6 & 7.2 & 0.0 & 6.2 & 2.7 & 7.1 & 3.5 & 1.9 & 1.1 & 4.2 \\
\multicolumn{1}{c|}{} &  & Multi. & 4.3 & 6.8 & 7.9 & 1.5 & 0.8 & 12.9 & 3.0 & 0.6 & 7.2 & 0.0 & 6.5 & 2.9 & 8.7 & 3.7 & 2.3 & 1.3 & 4.4 \\ \cline{2-20}

\multicolumn{1}{c|}{} & \multirow{2}{*}{S2U} & Mono. & 0.8 & 5.1 & 5.4 & 0.0 & 0.0 & 6.8 & 0.0 & 0.0 & 1.6 & 0.0 & 7.0 & 0.0 & 0.0 & 0.0 & 0.0 & 0.0 & 1.7 \\
\multicolumn{1}{c|}{} & & Multi. & 0.8 & 5.3 & 5.4 & 0.0 & 0.0 & 7.1 & 0.0 & 0.0 & 1.5 & 0.0 & 7.3 & 0.0 & 0.0 & 0.0 & 0.0 & 0.0 & 1.7 \\ \cline{2-20}
\multicolumn{1}{c|}{} & Textless & Mono. & 2.7 & 2.7 & 6.0 & 0.7 & 0.6 & 10.4 & 2.4 & 0.3 & 3.6 & 0.1 & 3.8 & 1.3 & 5.1 & 2.0 & 1.2 & 1.2 & 2.8 \\ \thickhline
\end{tabular}}
\end{table*}

The other bilingual baseline is Textless model \cite{DBLP:journals/corr/abs-2211-04508}, which uses the same training and validation data as models above. In Textless model, the speech encoder has $2$ convolutional layers and $12$ Transformer encoder layers with $512$-d layer and $2048$-d forward embeddings. Its unit decoder consists of $6$ Transformer decoder layers with layer and forward embedding of $512$ and $2048$ dimensions. Textless model has $70$M parameters. 

\textbf{Hyperparameters}. The multilingual speech-to-masked-unit model has a dropout probability of $0.1$ and a label smoothing factor of $0.2$. It is trained with a batch size of $320$k tokens and a update frequency of $8$ on $48$ GPUs. The learning rate is set as $0.0001$. The total number of training steps is $200$k, and it takes $10$ days to train a dense multilingual model. The best checkpoints which have the lowest loss on the validation set are used for S2ST evaluation. The bilingual speech-to-unit models have the same hyperparameters as the multilingual speech-to-masked unit model, and they are trained for $50$k steps, which takes around $2$ days. The checkpoints with the best validation loss are evaluated.

\textbf{Evaluation}. We have in-domain and out-of-domain test data for S2ST evaluation \cite{DBLP:journals/corr/abs-2211-04508}. The in-domain testsets are collected from EuroParl-ST (EP) and VoxPopuli (VP) corpus, whose European Parliament speech is in the same domain as our training data. FLEURS serves as out-of-domain data, and its test split is taken as test data. Following previous works on S2ST, we report ASR-BLEU as the metric of translation quality. The generated waveform by models are transcribed by prerained ASR models, and then BLEU score is calculated by comparing the transcriptions with reference target texts.

\textbf{Results}. Monolingual vocoders used in S2ST experiments are ``Mono-S'' and multilingual vocoders are ``Multi-L (+LID)'' as described in \autoref{exp:vocoder}. \autoref{tab:s2s_bleu} reports ASR-BLEU of S2ST models on testsets. Bilingual S2U models perform well in high-resource directions such as en-es and en-nl, but have low BLEU in other directions. The large S2U model is known to be data-hungry and are not trained well in languages without sufficient data. As for bilingual Textless models which have fewer parameters, they fall behind S2U models in high-resource languages, but outperform S2U in low-resource directions.

S2MU model together with multilingual vocoder achieves the best performance. The average gains over bilingual S2U with monolingual vocoder are +$5.2$ and +$2.7$ BLEU on in-domain and out-of-domain testsets respectively. When compared with bilingual Textless models, the average gains are +$1.1$ and +$1.6$ BLEU. 

With S2MU model, multilingual vocoders outperforms monolingual vocoders by $0.5$ and $0.2$ BLEU averaged over $16$ directions on in-domain and out-of-domain data respectively. Looking at each language direction, the BLEU gain on S2ST by multilingual vocoder is correlated with WER reduction on resynthesis. BLEU gain is also dependent on inference performance of S2MU. For example, multilingual vocoder reduces WER of Slovak (sk) by $5\%$, but does not show much translation gains due to low-quality units. 

\subsection{Analysis}

Multilingual S2MU outperforms Textless models in directions except for three Slavic languages: hr, sk and sl. These three languages have very limited training data, and multilingual training is in favor of higher-resource languages. Even with more capacity in S2MU, these languages don't benefit from multilingual training.

When we compare bilingual models, S2U and Textless, we find that model capacity should match the language resource size in order to optimize translation performance. Given high-resource languages including it and de, S2U with much more parameters demonstrate gains over smaller Textless model. As for languages with less training data, the performance of S2U drops sharply and BLEU scores are close $0$ in pt and ro, while Textless model achieves higher BLEU of $11.8$ and $7.6$ respectively. This suggests that model capacity is a bottleneck if there is sufficient data, while data size becomes the blocker if the model is too large.

When it comes to extremely low-resource directions such as et, fi and lt, all models perform poorly. We also note that data domain matters to the translation performance. For each model, its performance is always better on in-domain sets than on out-of-domain data.

Comparing multilingual vocoders against monolingual vocoders, the gains are more obvious in language directions with high BLEU scores. For example with S2MU model, multilingual vocoder improves BLEU by $3.9$ and $1.6$ on EP/VP and FLEURS data respectively. As for en-es translation on EP/VP testsets, multilingual vocoder brings +$2.0$ BLEU with S2MU model, and +$1.7$ BLEU with S2U model.


\section{Limitations}

This work proposes multilingual training techniques for speech-to-speech translation into multiple target languages. There have been extensive studies on multilinguality in tasks of machine translation and language models, which could be leveraged to further improve multilingual S2ST. In our future work, we would like to explore more research ideas such as multilingual data sampling to deal with imbalanced training data. Also this work groups languages based on their linguistic similarity. According to findings of existing literature, a better grouping could be learned with a data-driven approach to encourage cross-lingual transfer and mitigate language interference. 

Furthermore, we have to concatenate multiple sets of units as the vocabulary due to HuBERT models trained for only one language family. It is worth exploring a feature extraction model (e.g. HuBERT) supporting all languages so that we could use a single unit vocabulary shared by all languages. The shared vocabulary might better support the knowledge transfer across languages, especially those from different families.

\section{Conclusions}

We developed a single multilingual model to support speech-to-speech translation into multiple target languages. We proposed vocabulary masking and multilingual vocoding to encourage knowledge transfer across languages and mitigate their interference at the same time. Empirical results demonstrated that these are useful techniques for multilingual S2S training. 




\bibliographystyle{IEEEtran}
\bibliography{mybib}

~

\appendix

\section{Empirical Details}

\subsection{ASR Models}

\autoref{tab:asr} reports the ASR models we used in the evaluation of vocoder and speech-to-speech translation.

\begin{table*}[htbp!]
\caption{ASR model identifiers on HuggingFace for each language.}
\label{tab:asr}
\resizebox{1.0\textwidth}{!}{
\begin{tabular}{cc|cc}
\thickhline
\textbf{Lang} & \textbf{ASR Model} & \textbf{Lang} & \textbf{ASR Model}\\ \thickhline
\textbf{cs} & comodoro/wav2vec2-xls-r-300m-cs-250 &  \textbf{de} & jonatasgrosman/wav2vec2-xls-r-1b-german \\ \hline
\textbf{et} & RASMUS/wav2vec2-xlsr-1b-et & \textbf{fi}  & jonatasgrosman/wav2vec2-large-xlsr-53-finnish \\ \hline
\textbf{hr} & classla/wav2vec2-xls-r-parlaspeech-hr & \textbf{hu} & jonatasgrosman/wav2vec2-large-xlsr-53-hungarian \\ \hline
\textbf{it} & jonatasgrosman/wav2vec2-large-xlsr-53-italian & \textbf{lt} & sammy786/wav2vec2-xlsr-lithuanian \\ \hline
\textbf{nl} & jonatasgrosman/wav2vec2-xls-r-1b-dutch & \textbf{pl} & jonatasgrosman/wav2vec2-xls-r-1b-polish \\ \hline
\textbf{pt} & jonatasgrosman/wav2vec2-xls-r-1b-portuguese & \textbf{ro} & gigant/romanian-wav2vec2 \\ \hline
\textbf{sk} & anuragshas/wav2vec2-xls-r-300m-sk-cv8-with-lm  & \textbf{sl} & anuragshas/wav2vec2-xls-r-300m-sl-cv8-with-lm \\ \thickhline
\end{tabular}}
\end{table*}

\subsection{Vocoder Resynthesis}

\textbf{Language v. family units }. We reuse the data that has been used for vocoder training in SpeechMatrix \cite{DBLP:journals/corr/abs-2211-04508}. \autoref{tab:vocoder_data} provides details regarding data source and train set size in each language. The column ``Mono. (lang units)'' reports WER achieved by monolingual vocoders trained with language-specific units, which are extracted by k-means model trained on a specific language. These numbers are reported in SpeechMatrix \cite{DBLP:journals/corr/abs-2211-04508}. 

We are training multilingual vocoders by allowing multiple languages to share the same k-means model and have family-specific units. For a fair comparison with multilingual vocoders, we also train monolingual vocoders with family-specific units. Their WER is reported in the column ``Mono. (fam units)'' in \autoref{tab:vocoder_data}. As is shown, monolingual vocoders have comparable performance in speech resynthesis no matter whether the unit vocabulary is specific to languages or shared by a language family.

\begin{table}[htbp!]
\caption{WER ($\downarrow$) of resynthesized speech from monolingual vocoder trained with language- and family-specific units.}
\label{tab:vocoder_data}
\centering
\resizebox{0.48\textwidth}{!}{
\begin{tabular}{cccccc}
\thickhline
\begin{tabular}[c]{@{}c@{}}Family\\~ \end{tabular} & \begin{tabular}[c]{@{}c@{}}Language\\~ \end{tabular} & \begin{tabular}[c]{@{}c@{}}Source \\~ \end{tabular} & \begin{tabular}[c]{@{}c@{}}Train \\ hours\end{tabular} & \begin{tabular}[c]{@{}c@{}}Mono.\\ (lang units)\end{tabular} & \begin{tabular}[c]{@{}c@{}}Mono.\\ (fam units)\end{tabular} \\ \hline
\multirow{2}{*}{Gem} & de & CSS10 & 13.2 & 16.1 & 16.1 \\
 & nl & CSS10 & 11.3 & 27.0 & 28.0 \\ \hline
\multirow{5}{*}{Rom} & es & CSS10 & 23.4 & 12.0 & 11.3 \\
 & fr & CSS10 & 17.7 & 29.3 & 30.7 \\
 & it & VP & 25.8 & 27.4 & 31.6 \\
 & pt & CV & 16.1 & 31.1 & 36.6 \\
 & ro & VP & 25.5 & 50.4 & 50.0 \\ \hline
\multirow{5}{*}{Slavic}  & cs & VP & 26.8 & 23.0 & 23.0 \\
 & hr & VP & 25.3 & 29.0 & 29.7 \\
 & lt & VP & 1.3 & 57.3 & 57.3 \\
 & pl & VP & 26.7 & 23.2 & 25.0 \\
 & sk & VP & 25.3 & 40.7 & 41.0 \\
 & sl & VP & 6.1 & 46.3 & 47.0 \\ \hline
\multirow{3}{*}{Uralic} & et & CV & 12.0 & 44.3 & 44.1 \\
 & fi & CSS10 & 8.3 & 14.6 & 17.8 \\
 & hu & CSS10 & 7.9 & 21.3 & 21.0 \\ \thickhline
\end{tabular}}
\end{table}

\subsection{Speech-to-Speech Translation}

\end{document}